\documentclass{article}
\usepackage{spconf,amsmath,graphicx}
\usepackage{amsfonts}
\usepackage{booktabs}
\usepackage{multirow}
\usepackage{bm}
\usepackage{hyperref}
\hypersetup{hidelinks}
\usepackage{color}
\usepackage{float}
\usepackage{stmaryrd}
\usepackage{algorithm}
\usepackage{algorithmic}
\usepackage{hyperref}
\usepackage{caption}
\usepackage{subcaption}
\usepackage{flushend}

\newcommand{\norm}[1]{\left\lVert \, #1 \, \right\rVert}

\newcommand{\KOp}[1]{\llbracket #1 \rrbracket} 

\newcommand{\vect}[1]{\mathbf{#1}}

\newcommand{\vc}{\mathbf{c}}

\newcommand{\vx}{\mathbf{x}}

\newcommand{\vmu}{\boldsymbol{\mu}}

\newcommand{\matr}[1]{\mathbf{#1}}
\newcommand{\mA}{\matr{A}}
\newcommand{\mB}{\matr{B}}
\newcommand{\mC}{\matr{C}}
\newcommand{\mD}{\matr{D}}
\newcommand{\mE}{\matr{E}}
\newcommand{\mF}{\matr{F}}
\newcommand{\mG}{\matr{G}}
\newcommand{\mH}{\matr{H}}
\newcommand{\mI}{\matr{I}}

\newcommand{\mX}{\matr{X}}
\newcommand{\mY}{\matr{Y}}
\newcommand{\mZ}{\matr{Z}}

\newcommand{\mDelta}{\matr{\Delta}}

\newcommand{\R}{\mathbb{R}}

\makeatletter
\newcommand{\argmin}{\mathop{\text{~argmin~}}}

\DeclareMathOperator{\prox}{prox}
\DeclareMathOperator{\vecn}{vec}

\newcommand{\tens}[1]{\boldsymbol{\mathcal{#1}}}

\newcommand{\tX}{\tens{X}}
\newcommand{\tY}{\tens{Y}}
\newcommand{\tZ}{\tens{Z}}

\newcommand{\tN}{\tens{N}}

\title{PARAFAC2-based Coupled Matrix and Tensor Factorizations}
\name{Carla Schenker$^{1,2,\dagger}$\qquad Xiulin Wang$^{3,4,1\dagger}$ \qquad Evrim Acar$^{1}$\thanks{$^{\dagger}$These authors contributed equally to the work.}}

\address{
\small{$^{1}$ Simula Metropolitan Center for Digital Engineering, Oslo, Norway,
$^{2}$ Oslo Metropolitan University, Oslo, Norway}\\
\small{$^{3}$ Department of Radiology, Affiliated Zhongshan Hospital of Dalian University, Dalian,
China} \\
\small{$^{4}$ School of Biomedical Engineering, Dalian University of Technology, Dalian, China} \\
}
\begin{document}
%
\maketitle
\begin{abstract}
Coupled matrix and tensor factorizations (CMTF) have emerged as an effective data fusion tool to jointly analyze data sets in the form of matrices and higher-order tensors. The PARAFAC2 model has shown to be a promising alternative to the CANDECOMP/PARAFAC (CP) tensor model due to its flexibility and capability to handle irregular/ragged tensors. While fusion models based on a PARAFAC2 model coupled with matrix/tensor decompositions have been recently studied, they are limited in terms of possible regularizations and/or types of coupling between data sets. In this paper, we propose an algorithmic framework for fitting PARAFAC2-based CMTF models with the possibility of imposing various constraints on all modes and linear couplings, using Alternating Optimization (AO) and the Alternating Direction Method of Multipliers (ADMM). Through numerical experiments, we demonstrate that the proposed algorithmic approach accurately recovers the underlying patterns using various constraints and linear couplings.

\end{abstract}
\section{Introduction}
\label{sec:intro}
Joint analysis of heterogeneous data from multiple sources has the potential to capture complementary information and reveal underlying patterns of interest. Coupled matrix and tensor factorizations (CMTF) are an effective approach to jointly analyze such data in the form of matrices and tensors in various fields, e.g., social network analysis \cite{AcKoDu11b, ErAcCe13, Araujo2019Tensorcast}, neuroscience \cite{chatzichristos2018fusion, AcShLe19, wang2020group}, bioinformatics \cite{TaMu21} and remote sensing \cite{KaFuSi18}. CMTF models approximate each dataset using a low-rank model, where some factors/patterns are shared between data sets. Couplings with (linear) transformations have proven useful in many applications, \textit{e.g.,} accounting for different spatial, temporal or spectral relations between datasets \cite{chatzichristos2018fusion,KaFuSi18}, or modeling partially shared components \cite{chatzichristos2018fusion,AcShLe19}.
For analyzing higher-order tensors, CMTF methods often rely on the CANDECOMP/PARAFAC (CP) model \cite{harshman1970foundations,carroll1970analysis}, which approximates the tensor as a sum of
rank-one tensors. However, the CP model has strict multilinearity assumptions and cannot handle irregular tensors. 
The PARAFAC2 model \cite{Hars1972b, KiTeBr99} relaxes the CP model by allowing one factor matrix to vary across tensor slices and enables the decomposition of irregular tensors. 
PARAFAC2 has shown to be advantageous in chromatographic data analysis (with unaligned profiles) \cite{Bro1999PARAFAC2PartIM}, temporal phenotyping (with unaligned clinical visits) \cite{PePa19}, and tracing evolving patterns \cite{roald2020tracing}.

Recent CMTF studies have incorporated the PARAFAC2 model. For instance, Afshar et al. \cite{afshar2020taste} use a non-negative PARAFAC2 model coupled with a non-negative matrix factorization to jointly analyze electronic health records 
and patient demographic data. In \cite{chatzichristos2018fusion}, linearly coupled tensor decompositions are used to jointly analyze neuroimaging signals from different modalities, where PARAFAC2 is used to cope with subject variability. However, previous studies have been limited in terms of constraints on the factors and/or different types of couplings between data sets. Usually, the factor matrix of the varying mode in PARAFAC2 is estimated implicitly \cite{chatzichristos2018fusion, KiTeBr99}, which makes it challenging to impose constraints. The TASTE framework \cite{afshar2020taste}, therefore, adapts a $\textit{flexible}$ PARAFAC2 constraint \cite{cohen18flexible}, which allows for non-negativity constraints on the varying mode. TASTE has also been generalized to the coupling of a PARAFAC2 model with a CP model together with different options for solving the (non-negative) least-squares sub-problems \cite{guiral2020c}. Still, this framework is limited to the unconstrained and non-negative case, and does not support partial- or other linear couplings.  

\begin{figure}[t!]
\centering
\includegraphics[width=\columnwidth]{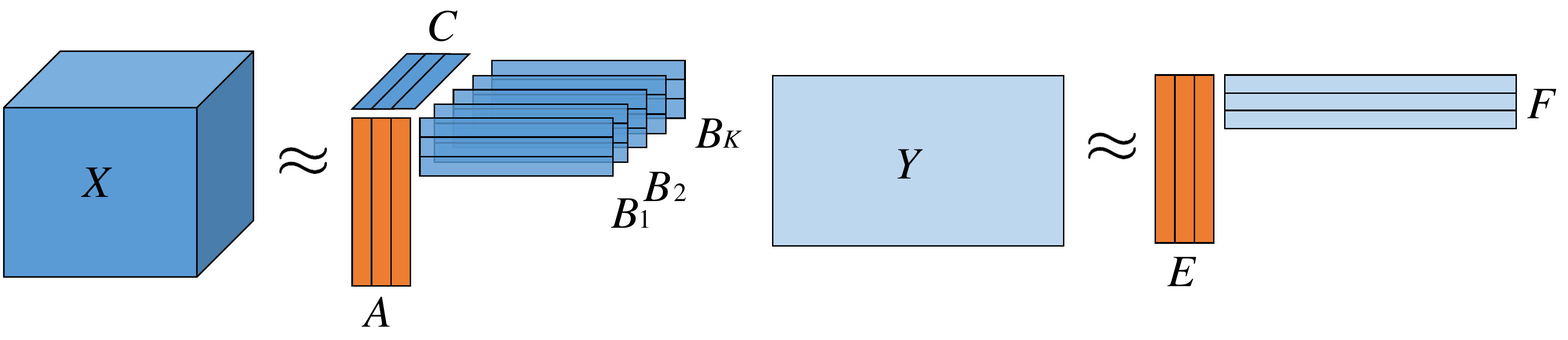}
\vspace*{-5mm}\caption{\small{A PARAFAC2 model coupled with a matrix factorization.}}
\label{fig:syn1B}
\end{figure}

In this paper, we introduce an AO-ADMM-based algorithmic approach for CMTF models incorporating the PARAFAC2 model referred to as \emph{PARAFAC2-based CMTF}. The framework accommodates linear couplings with (multiple) matrix- or CP-decompositions (Fig. \ref{fig:syn1B} and \ref{fig:syn3B}), and a variety of possible constraints and regularizations on all modes. Our algorithmic approach builds onto the AO-ADMM algorithm \cite{roald2022aoadmm} for constrained PARAFAC2, which allows for any proximal constraint in any mode, and the flexible framework for CP-based CMTF \cite{schenker2020flexible}. Using numerical experiments, we demonstrate the flexibility and accuracy of the proposed approach with different constraints 
and linear couplings. Furthermore, we show the promise of PARAFAC2-based CMTF models in terms of jointly analyzing dynamic and static data.
\section{PARAFAC2-based CMTF}
\label{sec:methods}
\begin{figure}[t!]
\centering
\includegraphics[width=\columnwidth]{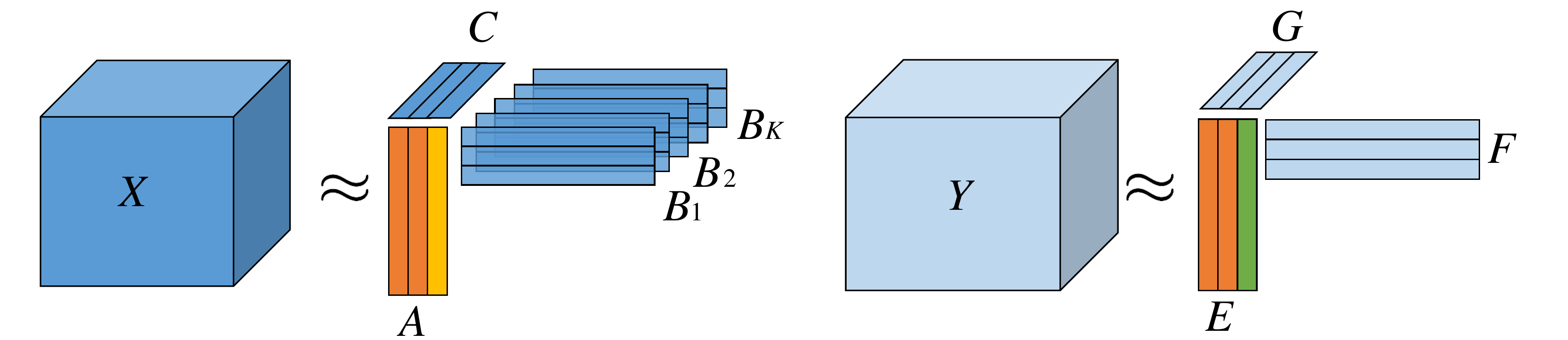}
\vspace*{-5mm}\caption{\small{A PARAFAC2 model partially coupled with a CP model.}}
\label{fig:syn3B}
\end{figure}

We are interested in CMTF models, where a PARA\-FAC2 model is coupled with a matrix- and/or CP-decomposition in one or several modes (Fig. \ref{fig:syn1B}, \ref{fig:syn3B}). In this paper, we focus on coupling in the first mode $\mA$ only. For the sake of readability, we consider PARAFAC2 coupled with a matrix decomposition here, while PARAFAC2 coupled with a CP model is considered in Sec. \ref{sec:experiment3}.
The PARAFAC2 model approximates the slices of a (ragged) tensor $\tX$, $\mX_k \in \R^{I_1\times J_k}$, $k\leq K$, with a low-rank factorization of rank $R_1$ as follows \cite{Hars1972b}:
 \begin{equation*}
 \small
\mX_k \approx \mA \mD_k \mB_k^T, \ \ \ \mB_k \in \mathcal{P},  \ \ \ \text{for} \ \  k=1,...,K,
\end{equation*}
where $\mD_k \in \R^{R_1 \times R_1}$ is diagonal, $\mA \in \R^{I_1 \times R_1}$ and $\mB_k \in \R^{J_k \times R_1}$. $\mC$ denotes the matrix that contains the diagonals of $\mD_k$ as rows, $\mD_k=\text{Diag}(\vc_{k,:})$. Unlike the CP model, where $\mX_k \approx \mA \mD_k \mB^T$, PARAFAC2 allows for the patterns in $\mB_k$ to vary across one mode. The set 
$\small{\mathcal{P} = \left\lbrace\left\lbrace \mB_k \right\rbrace_{k=1}^K | \mB_{k_1}^T\mB_{k_1}=\mB_{k_2}^T\mB_{k_2} \forall k_1,k_2\leq K \right\rbrace}$ defines the constant cross-product constraint of PARAFAC2, which ensures a unique decomposition (up to scaling and permutation ambiguities) 
under certain conditions \cite{KiTeBr99}.
The additional data matrix ${\mY}\in\mathbb{R}^{I_2\times L}$ is factorized using $R_2$ components as ${\mY} \approx {\mE \mF}^{T}$, with ${\mE}\in\mathbb{R}^{I_2\times R_2}$ and ${\mF}\in\mathbb{R}^{L\times R_2}$, such that matrices $\mA$ and $\mE$ are linearly coupled, \textit{i.e.,} the coupling can be written as $ \mH_{\mA} \vecn(\mA) = \mH_{\mA}^{\Delta} \vecn(\mDelta)$ and $\mH_{\mE} \vecn(\mE) = \mH_{\mE}^{\Delta} \vecn(\mDelta)$ with some unknown generating variable $\mDelta$ and known transformation matrices $\mH_{\mA,\mE},\mH_{\mA,\mE}^{\Delta}$ \cite{schenker2020flexible}. 
Using these transformations, we are able to model, for instance, averaging, blurring and downsampling as in \cite{KaFuSi18}, or convolution as in \cite{chatzichristos2018fusion}, as well as partially shared components. 
Using a Frobenius norm loss, the coupled factorization problem is formulated as follows, 
\begin{equation}\label{eq:costfunc}\small{
\begin{aligned}
    & \underset{\underset{
       \mA,\mE,\mF,\mDelta}{\left\lbrace\mD_k, \mB_k\right\rbrace_{k}}}{\argmin} & & 
  \!\sum\limits_{k=1}^{K}\left[w_1 \!\norm{\!\mX_k \!-\! \mA \mD_k \mB_k^T\!}_F^2 \!+\! g_D(\!\mD_k\!) \!+\! g_B(\!\mB_k\!)\right]\\
   & & & \ \!+\! w_2 \norm{\mY \!-\! \mE \mF^T}_F^2 \!+\! g_A(\mA) \!+\! g_E(\mE) \!+\! g_F(\mF) \\
   & \ \ \ \ \ \ \ \ \text{s.t. } & & \left\lbrace\mB_k\right\rbrace_{k\leq K} \in \mathcal{P},\\
   & & & \!\mH_{\!\mA}\! \vecn(\!\mA\!)\! =\! \mH_{\!\mA}^{\!\Delta}\! \vecn(\!\mDelta\!)\!,\!\mH_{\mE}\! \vecn(\mE)\! =\! \mH_{\!\mE}^{\!\Delta}\! \vecn(\!\mDelta\!),
\end{aligned}}
\end{equation}
where $g$s are regularization functions. This includes hard constraints via characteristic functions, \textit{e.g.,} $g=\iota_{\R^+}$ for non-negativity. We only require that proximal operators of the regularization functions are computable. \\
\noindent {\bf Algorithm.} We solve \eqref{eq:costfunc} using Alternating Optimization (AO) over the modes, \textit{i.e.,} alternatingly, all factor matrices of one mode (across tensors) are updated, while factor matrices of all other modes are kept constant. Subproblems for each mode are convex and solved using ADMM \cite{BoPaCh11}. This results in independent ADMM updates for uncoupled factor matrices 
and a joint update for coupled factor matrices such as $\mA$ and $\mE$ in \eqref{eq:costfunc}. 
Defining split variables $\mZ_A$ and $\mZ_E$, the subproblem for regularized $\mA$ and $\mE$ is written as:
\begin{equation}\small{
\begin{aligned}
    & \underset{\underset{\mZ_E, \mZ_A, \mDelta}{
       \mE, \mA}}{\argmin} & &
 w_1 \sum\limits_{k=1}^{K} \norm{\mX_k \!- \!\mA \mD_k \mB_k^T}_F^2 \!+\!  w_2\norm{\mY\! -\! \mE \mF^T}_F^2\\
 & & & \ \ \ \ \ \ \ \ \ \ \ \ \ \ \ \ \ \ \ \ \ \ \ \ \ \ \ \ \ \ \ \ \ + g_A(\mZ_A) + g_E(\mZ_E)   \\
   & \ \ \ \ \ \ \ \ \text{s.t. } & &
    \mA = \mZ_A , \ \ \mE = \mZ_E \\
    & & & \!\mH_{\!\mA}\! \vecn(\!\mA\!)\! =\! \mH_{\!\mA}^{\!\Delta}\! \vecn(\!\mDelta\!),\mH_{\mE}\! \vecn(\mE)\! =\! \mH_{\!\mE}^{\!\Delta}\! \vecn(\!\mDelta\!).
\end{aligned}}
\end{equation}
Interpreting $\mDelta$ as another split variable, ADMM is used to solve this problem. The solution of subproblems at factor matrix level depends on the structure of linear coupling. In \cite{schenker2020flexible}, we derive efficient updates (in the context of CP-based CMTF) for five types of linear coupling. For brevity, here we give the updates for exact coupling, \textit{i.e.,} $\mA=\mDelta$. In this case, $\mA$ and $\mE$ are updated by solving the following linear systems, 
\begin{equation}\label{eq:A_update}
\small{\begin{aligned} &\mA^{(n+1)} \left[w_1 \sum\limits_{k=1}^{K} \mD_k \mB_k^T \mB_k \mD_k\! +\!  \frac{\rho_A}{2} \left(\mI_{R_1}\! +\! \mI_{R_1} \right) \right] =\\
	            		  &\left[ w_1\sum\limits_{k=1}^{K}\mX_k \mB_k \mD_k \!+\! \frac{\rho_A}{2}\left( \mZ_{A}^{(n)}\!-\!\vmu_{Z_{A}}^{(n)} \!+\! \mDelta^{(n)}\! -\! \vmu_{\mDelta_A}^{(n)}\right)\right],
\end{aligned}}		  
\end{equation}
\begin{equation}\label{eq:E_update}
\small{\begin{aligned} &\mE^{(n+1)} \left[w_2 \mF^T \mF +  \frac{\rho_E}{2} \left(\mI_{R_2} + \mI_{R_2} \right) \right] =\\
	            		  &\left[ w_2 \mY \mF + \frac{\rho_E}{2}\left( \mZ_{E}^{(n)}-\vmu_{Z_{E}}^{(n)} + \mDelta^{(n)} - \vmu_{\mDelta_E}^{(n)}\right)\right],
\end{aligned}}
\end{equation}
where $\vmu_{Z_{A}}$, $\vmu_{Z_{E}}$, $\vmu_{\mDelta_A}$ and $\vmu_{\mDelta_E}$ are dual variables, $\rho_A$, $\rho_E$ are step sizes (computed as in \cite{roald2022aoadmm,schenker2020flexible}), and $\mI_{R}$ is an $R \times R$ identity matrix. The update for $\mDelta$ is in this case as follows:
\begin{equation}\label{eq:Delta_update}
\small{\begin{aligned}\!\mDelta^{(n\!+\!1)} \!=\! \tfrac{1}{\rho_A\!+\!\rho_E}\!\left[\! \rho_A \left(\!\mA^{(n\!+\!1)}\!+\!\vmu_{\mDelta_A}^{(n)} \!\right)\!+\!\rho_E \left(\!\mE^{(n\!+\!1)}\!+\!\vmu_{\mDelta_E}^{(n)}\! \right)\!\right]. \end{aligned}}
\end{equation}
The whole ADMM algorithm for this subproblem is given in Alg. \ref{alg:admm_Amode}, where 
$\small{\prox_{\frac{1}{\rho_A}g_{A}}(\vx)=\underset{\vect{u}}{\argmin} g_A(\vect{u}) + \frac{\rho_A}{2} \norm{ \vx - \vect{u}}_2^2}$
denotes the proximal operator of $g_{A}$. Updates for other types of coupling can be derived based on \eqref{eq:costfunc} and \cite{schenker2020flexible}, and are provided in the supplementary\footnote{\label{Git}\url{https://github.com/AOADMM-DataFusionFramework}} together with updates for other modes, that have been studied previously \cite{roald2022aoadmm,schenker2020flexible}. Since $\mB_k$ matrices are not coupled, their update is the same as in \cite{roald2022aoadmm}, using ADMM with an extra splitting variable for the constraint $\mB_k\! \in\! \mathcal{P}$. 
In practice, the proposed framework can handle any number of coupled tensors (CP/PARAFAC2) and/or matrices, and our code is publicly available$^{\ref{Git}}$. 
\begin{algorithm}[h!]

        \begin{algorithmic}[1]
            \WHILE{convergence criterion is not met}

	            \STATE{$\mA^{(n+1)} \longleftarrow \text{solve linear system  } \eqref{eq:A_update}
	            		  $}
	            		 
	            \STATE{$\mE^{(n+1)} \longleftarrow \text{solve linear system  } \eqref{eq:E_update}$}
	            		 
\STATE{$\mDelta^{(n+1)} \longleftarrow \text{  } \eqref{eq:Delta_update}$}	            		

				\STATE{$\small{\begin{aligned} \mZ_{A}^{(n+1)}=
		            \prox_{\frac{1}{\rho_A}g_{A}}\left(\mA^{(n+1)}+\vmu_{Z_{A}}^{(n)}  \right)\end{aligned}} $}
		         \STATE{$\small{\begin{aligned} \mZ_{E}^{(n+1)}=
		            \prox_{\frac{1}{\rho_E}g_{E}}\left(\mE^{(n+1)}+\vmu_{Z_{E}}^{(n)}  \right)\end{aligned}} $}
	            
	            \STATE{$\small{ \vmu_{Z_{A}}^{(n+1)} = \vmu_{Z_{A}}^{(n)} +
	            \mA^{(n+1)} - \mZ_{A}^{(n+1)} }$}
	            
	            \STATE{$\small{ \vmu_{Z_{E}}^{(n+1)} = \vmu_{Z_{E}}^{(n)} +
	            \mE^{(n+1)} - \mZ_{E}^{(n+1)} }$}
	            
	            \STATE{$\small{ \vmu_{\mDelta_A}^{(n+1)} = \vmu_{\mDelta_A}^{(n)} +  \mA^{(n+1)} - \mDelta^{(n+1)} }$}
	            
	            \STATE{$\small{ \vmu_{\mDelta_E}^{(n+1)} = \vmu_{\mDelta_E}^{(n)} + \mE^{(n+1)} - \mDelta^{(n+1)} }$}

            \STATE{$\small{n = n+1}$}
            \ENDWHILE{}
        \end{algorithmic}
        \caption{ADMM for subproblem w.r.t. $\mA$ and $\mE$}
\label{alg:admm_Amode}
\end{algorithm}
\vspace*{-5mm}
\section{Experiments and Results}
\label{sec:simu_res}
Using experiments on simulated data, we demonstrate that the proposed algorithmic approach can accurately reveal the underlying factors in different settings with various constraints and linear couplings.

\noindent{\bf Experimental Set-up:} 
For each experiment, we generate 20 random dataset pairs $\tX$ and $\tY$, where $\tX$ follows a PARAFAC2 model and $\tY$ either a CP- or a matrix decomposition with known ground-truth factor matrices. Noise is added to each dataset as $\small{{\tX}_{\textrm{noisy}} = {\tX}+\eta(\|{\tX}\|_{F} / \|{\tN}\|_{F}){\tN}}$, 
where $\eta$ is the noise level, $\small{\tN}$ is a noise tensor with entries drawn from the standard normal distribution. Each dataset is normalized to Frobenius norm $1$ and weights $w_i$ are set to $0.5$. 
We use multiple random initializations, and report the results only for the run with the lowest function value. Factor matrices are initialized by drawing from the standard normal,~or, in the case of non-negative factors, uniform distribution, and columns are normalized. 
Stopping conditions are given in the supplementary.
The accuracy of recovered factor matrices is measured using the Factor Match Score (FMS) defined as
\begin{equation*}
\small
\textrm{FMS}_{\bf U}	=1/R	\sum_{r}|{\bf u}_r^{T}\tilde{\bf u}_r|/(\left\|{\bf u}_r\right\|_2\left\|\tilde{\bf u}_r\right\|_2),
\vspace{-3mm}
\end{equation*} 
where ${\bf u}_r$ and $\tilde{\bf u}_r$ correspond to the $r$th column of true factor matrix $\bf U$ and recovered matrix $\tilde{\bf U}$ (after finding the best permutation of columns). For FMS$_\mB$, we concatenate all ${\bf B}_k$s to form ${\bf B}\in \mathbb{R}^{\sum J_k \times R}$.
Additionally, the model fit is used for the reconstruction error, $
\small
\textrm{Fit}=100\times(1-\|{\tZ}-\tilde{\tZ}\|_{F}^2/\|\tZ\|_F^2)$,
where tensor $\tilde{\tZ}$ denotes the reconstructed version of $\tZ$.
\vspace*{-3mm}
\subsection{Experiment 1: Exact Coupling \& Non-negativity}
\label{ex:exp1}
Here, we fit a PARAFAC2 model to a third-order tensor and jointly factorize a matrix using exact coupling ${\mA}={\mE}$ (as in Sec. \ref{sec:methods} and Fig. \ref{fig:syn1B}). We generate datasets of size $40~\!\times~\!120~\!\times~\!50$ and $40\!\times\! 60$ using random ground-truth factor matrices of rank $R=3$ drawn from the standard uniform distribution. ${\bf B}_k$s follow the PARAFAC2 constraint, and the factor matrix $\mC$ is shifted $(+0.1)$ to avoid near-zero elements. We fit the model with non-negativity constraints on all modes. Tab. \ref{tab:exp1} shows average model fit and FMS values at different noise levels demonstrating that the proposed algorithm can successfully recover the true underlying patterns.
\begin{table}[t!] 
\centering
\small
\caption{\small{Average model fit and FMS values for experiment 1.}}
\vspace*{-3mm}
\label{tab:exp1}
\renewcommand{\arraystretch}{1.2}
\begin{tabular}{cccccccc} %
\toprule 
\multirow{2}{*}{Noise($\eta$)}& \multicolumn{2}{c}{Fit (\%)}&&\multicolumn{4}{c}{FMS}\\
\cmidrule{2-3}\cmidrule{5-8}
&PAR2&Matrix&&A/E&B&C&F\\  
\midrule %
0& 100& 100&&1& 1& 1&1\\   %
0.2&96.17&96.63&&0.99&0.98&0.99&0.99\\
0.5&80.04&82.82&&0.91&0.92&0.98&0.92\\
\bottomrule %
\end{tabular}
\vspace*{-4mm}
\end{table}
\vspace*{-3mm}
\subsection{Experiment 2: Fusion of dynamic and static data}
\begin{figure}[t]
\centering
\includegraphics[width=\columnwidth]{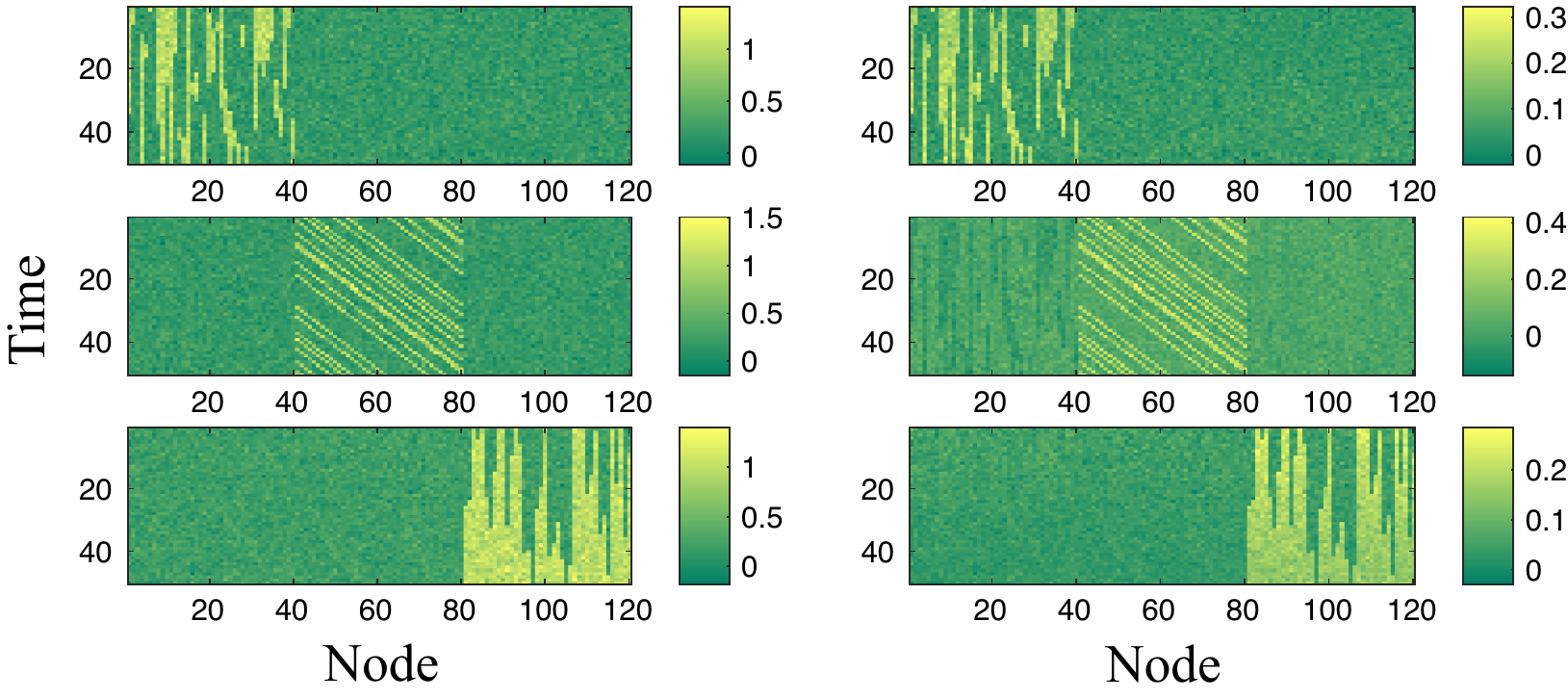}
\vspace*{-6mm}\caption{\small{Exp. 2: Ground-truth evolving networks ${\bf B}_k$s (left) and recovered ones (right) ($\textrm{Noise}=1, \textrm{ridge},\textrm{FMS}_{\mB}=0.973$).}}
\label{fig:syn2B}
\end{figure}

\begin{figure}[t]
\centering
\includegraphics[width=\columnwidth]{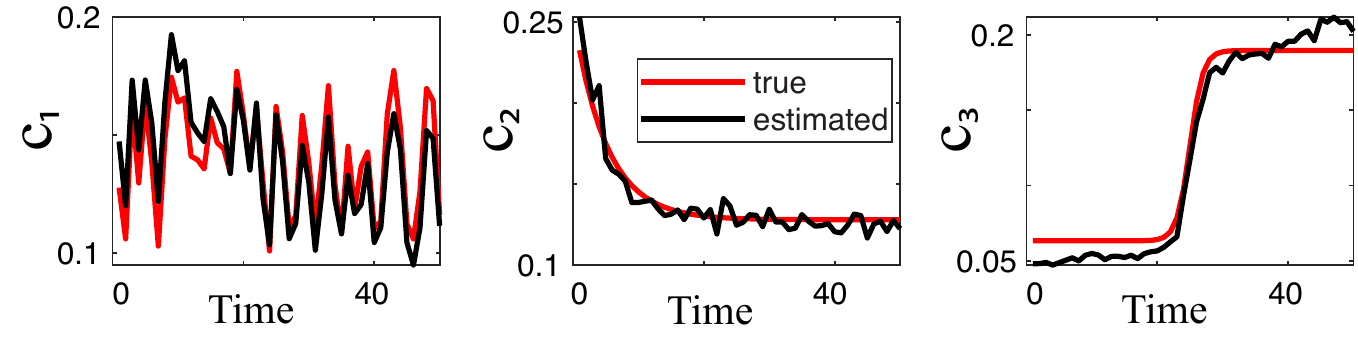}
\vspace*{-5mm}\caption{\small{Exp. 2: Ground-truth temporal patterns $\bf C$ (red) and recovered ones (black) ($\textrm{Noise}=1, \textrm{ridge}, \textrm{FMS}_{\mC}=0.997$).}}
\label{fig:syn2C}
\end{figure}

\begin{figure}[t]
\centering
\includegraphics[width=0.9\columnwidth]{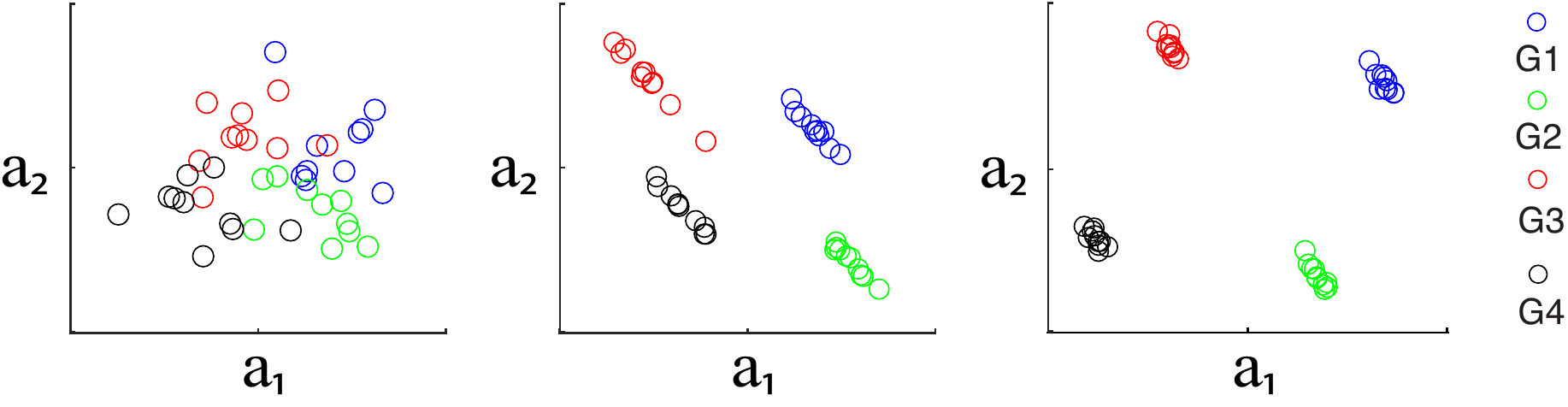}
\vspace*{-3mm}\caption{\small{Exp. 2: Example of clustering structure in the ground-truth $\mA$ (left: Noise=$1$) and recovered ones (middle: with coupling; right: with coupling and ridge) .}}
\label{fig:syn2A}
\end{figure}

In this example, we jointly analyze dynamic and static data sets using a PARAFAC2-based CMTF model where PARAFAC2 is used to capture evolving patterns from the dynamic data. 
When generating the data, ${\mB}_k$s do not follow the PARAFAC2 constraint, but are instead constructed to simulate networks that evolve along the temporal mode $\mC$, as in \cite{roald2020tracing}. As shown in Fig.~\ref{fig:syn2B}, ${\mB}_k$s have three columns ($R=3$), corresponding to a shrinking, shifting and growing network. $\mC$ is generated as a temporal pattern matrix that includes an exponential, a sigmoidal and a random curve (Fig.\ref{fig:syn2C}). In the coupled mode, a clustering structure with four clusters is embedded in the first two columns of $\mA$ and $\mE$. $\mF$ is generated as random non-negative. Dataset sizes are as in Experiment 1. No noise is added to the datasets; instead, matrix $\mA$ is perturbed by noise before constructing $\tX$. 
We compare the performance of models with and without exact coupling constraints between $\mA$ and $\mE$. We constrain $\mC$ and $\mF$ to be non-negative. Tab.~\ref{tab:exp2} shows the average performance, including clustering accuracy based on $k$-means clustering. Model fit and $\textrm{FMS}_{\mB}$ are never perfect for PARAFAC2 
since the true $\mB_k$s do not follow the PARAFAC2 constraint. Nevertheless, results show that the coupled model can capture the underlying patterns (including evolving networks) accurately (see also Fig. \ref{fig:syn2B} and \ref{fig:syn2C}) while also improving the clustering performance over the uncoupled case. 
We also 
show that ridge regularization (penalty $10^{-4}$) on all modes can improve the clustering performance for the noisy case, see Fig.~\ref{fig:syn2A}.

\begin{table*}[t] 
\centering
\caption{\small{Average performance for experiment 2. FMS of $\mA$ using true noisy $\mA$ (and using clean $\mA$ ($\mA=\mE$)).}}
\vspace*{-3mm}
\small
\label{tab:exp2}
\renewcommand{\arraystretch}{1.2}
\begin{tabular}{cccccccccccccc} %
\toprule 
\multirow{2}{*}{Ridge}&\multirow{2}{*}{Coupling}&\multirow{2}{*}{Noise}& \multicolumn{2}{c}{Fit (\%)}&&\multicolumn{5}{c}{FMS}&&\multicolumn{2}{c}{Clustering acc. (\%)}\\
\cmidrule{4-5}\cmidrule{7-11}\cmidrule{13-14}
&&&PAR2&Matrix&&A&B&C&E&F&&A&E\\  
\midrule %
\multirow{6}{*}{\text{no}}&\multirow{3}{*}{\text{no}}&0& 99.98& 100&&1 (1)&0.99& 1& -&-&&100&-\\   %
&&0.5&99.98&100&&1 (0.89)&0.99&1&-&-&&95.63&-\\
&&1&99.98&100&&1 (0.71)&0.99&1&-&-&&65.88&-\\
\cmidrule{2-14}
&\multirow{3}{*}{\text{yes}}&0& 99.75& 100&&1(1)&0.99& 0.99& 1&1&&100&100\\   %
&&0.5&84.51&99.98&&0.90(0.98)&0.99&1&0.98&0.99&&100&100\\
&&1&56.67&99.95&&0.72(0.96)&0.98&0.99&0.96&0.99&&99.63&99.63\\
\cmidrule{1-14}
\multirow{3}{*}{\text{yes}}&\multirow{3}{*}{\text{yes}}&0& 99.91& 100&&1(1)&0.99& 1& 1&1&&100&100\\   %
&&0.5&83.05&99.97&&0.90(0.99)&0.98&1&0.99&1&&100&100\\
&&1&57.53&99.95&&0.73(0.99)&0.96&1&0.99&1&&100&100\\
\bottomrule %
\end{tabular}
\end{table*}

\vspace*{-3mm}
\subsection{Experiment 3: Partial Coupling \& Smoothness}
\label{sec:experiment3}
\begin{figure}
\centering
\includegraphics[width=\columnwidth]{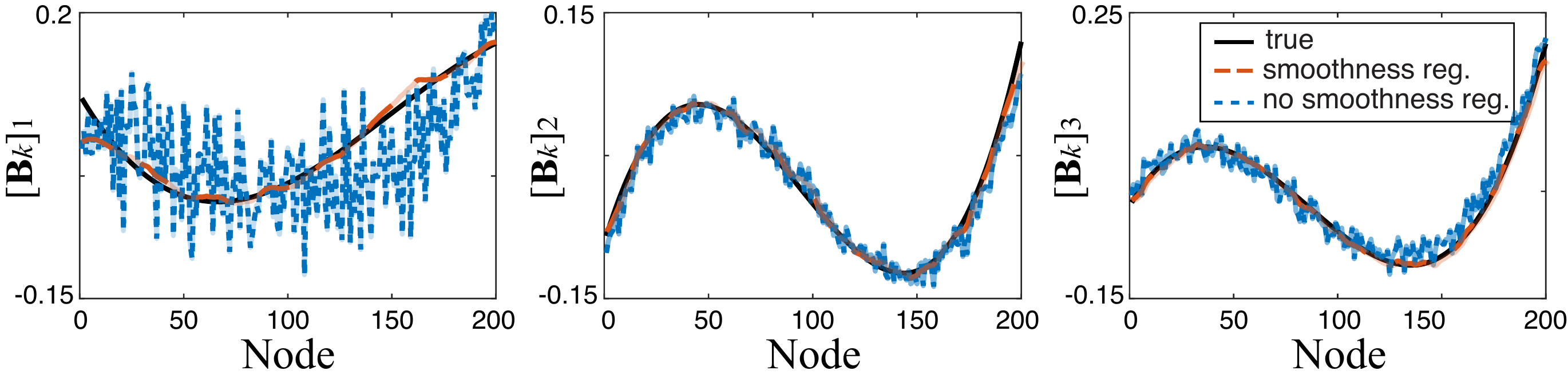}
\vspace*{-5mm}\caption{\small{Exp. 3: Example of recovery of the three components of $\mB_k$, with and without smoothness regularization.}}
\label{fig:syn3A}
\end{figure}

Here, we test our algorithm in a setting with partial coupling and smoothness regularization. We construct two tensors, one of size $30\! \times\! 200\! \times\! 30$ following a PARAFAC2 model, and another of size $30\! \times\! 20\! \times\! 50$ following a CP model. Both have three components, but only two components are shared in the first mode (Fig.\ref{fig:syn3B}). We generate smooth components in the $\mB_k$-mode as in \cite{roald2022aoadmm}. $\mC$ is generated as in Exp.$1$. Other factor matrices are generated from the standard normal distribution. 
The noise level is $0.5$. 
We then solve the following 
problem:
\vspace*{-2mm}
\begin{equation*}\small{
\begin{aligned}
    & \underset{\underset{
       \mA,\mE,\mF,\mDelta,}{\left\lbrace\mD_k, \mB_k\right\rbrace_{k\leq K}}}{\argmin} \frac{1}{2} \norm{\mY - \KOp{\mE, \mF,\mG}}_F^2 + \iota_{\mathcal{B}_1^2}(\mA) + \iota_{\mathcal{B}_1^2}(\mE) \\
    &\  +\sum\limits_{k=1}^{K}\left[\frac{1}{2} \norm{\mX_k  - \mA \mD_k \mB_k^T}_F^2 + \iota_{\mathcal{B}_{1,+}^2}(\mD_k) + g_\texttt{GL}(\mB_k)\right]\\
   & \ \ \ \  \text{s.t. } \ \ \ \ \ \ \ \ \ \ \   \left\lbrace\mB_k\right\rbrace_{k\leq K} \in \mathcal{P},
   \mA =  \mDelta \hat{\mH}_{\mA}^{\Delta}, \mE = \mDelta  \hat{\mH}_{\mE}^{\Delta}.
\end{aligned}}
\vspace*{-1mm}
\end{equation*}
Here, $g_\texttt{GL}$ denotes a columnwise graph laplacian regularization with strength $1$ to promote smooth components, see \cite{roald2022aoadmm}. 
In order to make the smoothness regularization effective, we constrain the factor vectors of modes $\mA,\mC$ and $\mE$ to be inside the unit $\ell 2$-ball $\mathcal{B}_1^2$. Mode $\mC$ is also constrained to be non-negative. We use linear coupling constraints 
to account for the partial coupling, \textit{i.e.,} $\hat{\mH}_{\mA}^{\Delta}$ 
indicates which columns from the ``dictionary" $\mDelta$ are present in $\mA$. Our algorithm is able to recover the true factors with FMS $0.99$ for mode $\mB$ and $1$ for all other modes, 
yielding smooth $\mB_k$ components (Fig.~\ref{fig:syn3A}). 

\section{Conclusion}
\label{sec:conl}
We have presented an AO-ADMM framework for fitting PARAFAC2-based CMTF models which supports coupling with matrix- and CP-decompositions, linear coupling and various constraints on every mode. Our experiments on synthetic datasets show that the algorithm reveals the underlying patterns accurately, even when the data does not exactly follow the PARAFAC2 constraint. We also show that such models can be useful in terms of jointly analyzing dynamic and static data by revealing evolving patterns and improving clustering performance, as well as facilitating interpretability through constraints other than non-negativity on the varying mode. 

\bibliographystyle{IEEEbib}
\bibliography{strings,refs}

\end{document}